\documentclass[conference]{IEEEtran}
\IEEEoverridecommandlockouts

\usepackage[nocompress]{cite}
\usepackage{amsmath,amssymb,amsfonts}
\usepackage{algorithmic}
\usepackage{graphicx}
\usepackage{textcomp}
\usepackage{xcolor}
\usepackage{booktabs}
\usepackage{multirow}
\def\BibTeX{{\rm B\kern-.05em{\sc i\kern-.025em b}\kern-.08em
    T\kern-.1667em\lower.7ex\hbox{E}\kern-.125emX}}

\begin{document}

\title{Gray-box Adversarial Attack of Deep Reinforcement Learning-based Trading Agents*\\
\thanks{This research was partly funded by NSERC Alliance – Alberta Innovates Advance Program: ALLRP/556396-2020 and 202102242.}
}

\author{\IEEEauthorblockN{1\textsuperscript{st} Foozhan Ataiefard}
\IEEEauthorblockA{\textit{Electrical and Software Engineering} \\
\textit{University of Calgary}\\
Calgary, Canada \\
foozhan.ataiefard1@ucalgary.ca}
\and
\IEEEauthorblockN{2\textsuperscript{nd} Hadi Hemmati}
\IEEEauthorblockA{\textit{Electrical Engineering and Computer Science} \\
\textit{York University}\\
Toronto, Canada \\
hemmati@yorku.ca}
}

\maketitle

\begin{abstract}
In recent years, deep reinforcement learning (Deep RL) has been successfully implemented as a smart agent in many systems such as complex games, self-driving cars, and chat-bots. 
One of the interesting use cases of Deep RL is its application as an automated stock trading agent.
In general, any automated trading agent is prone to manipulations by adversaries in the trading environment. Thus studying their robustness is vital for their success in practice. However, typical mechanism to study RL robustness, which is based on white-box gradient-based adversarial sample generation techniques (like FGSM), is obsolete for this use case, since the models are protected behind secure international exchange APIs, such as NASDAQ. 
In this research, we demonstrate that a ``gray-box'' approach for attacking a Deep RL-based trading agent is possible by trading in the same stock market, with no extra access to the trading agent. 
In our proposed approach, an adversary agent uses a hybrid Deep Neural Network as its policy consisting of Convolutional layers and fully-connected layers. 
On average, over three simulated trading market configurations, the adversary policy proposed in this research is able to reduce the reward values by 214.17\%, which results in 
reducing the potential profits of the baseline by 139.4\%, ensemble method by 93.7\%, and an automated trading software developed by our industrial partner by 85.5\%, while consuming significantly less budget than the victims (427.77\%, 187.16\%, and 66.97\%, respectively). 
\end{abstract}


\section{Introduction}
Application of deep neural networks in the field of automated trading has gained huge interest in recent years. Given the high capacity of DNNs to approximate complex and non-linear relations, their integration in reinforcement learning algorithms such as Q-learning has introduced a new family of solutions, i.e., Deep RL. Deep RLs have been successfully applied for control-based tasks such as video games \cite{mnih2013playing}, Go\cite{silver2016mastering}, automated driving in simulations and real world\cite{dosovitskiy2017carla}, and trading \cite{noonan2017jpmorgan}. Deep RLs in automated trading are a relatively new and under-studied topic. For instance, an ensemble method making decision from three different Deep RL algorithms has been proposed in one study \cite{yang2020deep} and an inverse reinforcement learning approach has been proposed in another \cite{roa2019adversarial}.

As efficient as these algorithms have been in solving complicated problems, they are still prone to adversarial perturbations to their inputs. Vision-based Deep RL policies have been shown to be vulnerable against adversarial examples, resulting in mis-classifications \cite{szegedy2013intriguing, huang2017adversarial}. In the prior studies on robustness of Deep RL agents, an attacking method has direct access to its victim's input. However, for many applications such as trading such access is considered to be almost infeasible. For vision-based agents, a study \cite{gleave2019adversarial} found that it's possible to find an adversarial policy that interacts with the victim's environment acting as another player.

Robustness to adversarial attacks are particularly important in a trading system since an adversary agent can legally act as trader but under the hood manipulate the market for a specific competitor or company/agent under attack. Thus the first step toward building robust Deep RL trader agents is to identify the weak points with respect to attacks, which first requires having a realistic and powerful adversarial sample generator. 

Therefore, in this paper, we propose a gray-box framework to create adversarial samples for Deep RL trading agents similar to trading in a real stock market. 
The gray-box assumption is that, the trading agents’ source code, policy architecture, DNN weights, and training algorithms are all unknown for the adversary. 
The only accessible data is the current state of the market and the decision of the trading agent (its chosen trading action in that given state, which is public in many trading platforms).
Our framework uses a real-time agent-based trading market simulation named ABIDES \cite{10.1145/3384441.3395986}. ABIDES is among the few open source trading simulations, which is capable of mimicking real stock markets and has been used in several studies published in financial venues. 

In order to show effectiveness of our adversary policy, we trained three trading agents using three most realistic configurations of the market in the simulator. After training these agents, they are integrated in a trading environment, where the adversary is allowed to trade as well. Three different aspects of the adversary are evaluated through three research questions (RQs), where we look at: (RQ1) how effective the proposed adversary is in changing the trader agent's decisions, (RQ2) to what extent it can change the trader’s profits, and (RQ3) can it do it within a reasonable cost, while staying systematic (i.e., indirectly affecting the victim's learnt policy)?

The contributions of this paper can be summarised as:
\begin{itemize}
  \item Providing an end-to-end solution to create gray-box adversarial attacks for Deep RL trader agents.
  \item Experimentally evaluating the attacks on three agents (including an industrial agent) and three market scenarios.
  \item Reporting evidence that the proposed approach can create  attacks, by a reasonable cost, which are successful in systematically (i.e., by affecting its learnt policy) changing the decisions of the trader for worse.
\end{itemize}

The replication package of this paper including the network architectures and hyper-parameters is publicly available ~\cite{repo}.

\section{Background}\label{bkg}

 
\subsection{Deep Reinforcement Learning for Trading} \label{DRLT}

The general optimization problem for stock market trading is definable as a Markov Decision Process, which is solvable using deep reinforcement learning algorithms. Optimization target of the RL agent is defined as maximizing profits. The elements of this RL problem are as follows:
\begin{itemize}
  \item State($s$): Vector containing agent's remaining balance, owned shares, current price of shares, best bid and ask prices for shares and technical indicators such as RSI.
  \item Action($a$): RL agent's choice of action according to the current state $s_t$. Actions for a trading agents can be buy, hold or sell a specific number of shares.
  \item Reward($r$): Reward function of RL agent for taking action $a$ at current environment state $s$:
  \[r(s,a,\hat{s}) \in \mathbb{R} \]
  \item Policy($\pi$): A Deep Neural Network mapping set of environment states $S$ to the set of possible actions $A$:
  \[\pi: S \rightarrow A \]
  
\end{itemize}

Most popular deep reinforcement learning algorithms employed in financial markets belong to one of categories of actor-critic, actor-only , critic-only approaches or an ensemble of these techniques \cite{fischer2018reinforcement}. 

Deep Q-learning based algorithms are the most common between critic-only approaches used for trading agents. In this group of algorithms, a deep neural network is trained to approximate a Q-value function. Q-value function tries to provide a close estimation of the expected reward for an action $a$ based on the current state $s$. The agent uses the Q-value to optimize a policy for choosing actions that are expected to return the maximum rewards in a given state. Actor-only algorithms are designed to work with discrete action spaces (i.e. buy, hold or neutral, sell), which limits the control over trading actions. 

Another family of popular algorithms for trading are Actor-only approaches, also called policy search approaches. These algorithms eliminate the need for predicting future rewards by learning best trading strategies directly from the environment. These algorithms use immediate rewards to optimize parameters of the policy. The policy itself, in essence is a probability distribution of actions representing a trading strategy. 

Most recent applications of deep reinforcement learning approaches in trading benefit from actor-critic approaches. In this category of RL algorithms two networks are simultaneously trained. First network learns the policy $\pi$ (actor) and Second network learns an estimation of value function $V^{\pi}(s)$ (critic). $V^{\pi}(s)$ predicts future rewards that will be received from the environment, starting from state $s$ and taking actions from $\pi$ network. To reach an efficient Policy, its network is updated using policy gradients according to $V$.

We test robustness of 3 automated trading RL agents using our adversary approach: Baseline agent, ensemble agent from \cite{yang2020deep} and an industrial agent as follows:


\subsubsection{Baseline Agent}
A typical actor-critic model with a two-headed fully-connected neural network. One head acting as policy output (action) and the other head is the value function. 

\subsubsection{Ensemble Agent}
A more sophisticated model consisting of three actor-critic algorithms. Each action is selected from the best performing agent among PPO, A2C and DDPG\cite{lillicrap2015continuous} algorithms. Both of these agents use the same reward function $R_t$ defined as below: 
\[R(s_t,a,s_{t+1}) = P_{t+1} - P_t\] 
$P_t$ or Portfolio value at time $t$ is total value of agent's assets including value of owned shares and cash balance. All of the agents in this study use the same state vector $S$, as defined. 

\subsubsection{Industrial Agent}
This is one of the agents developed by our industry partner that outperformed the above two agents. Although the overall architecture is similar to the ensemble agent it includes many detailed optimizations that due to confidentiality we can not reveal. The source code of all the available agents can be found in the replication package~\cite{repo}.


\subsection{Adversarial Policy in Deep Reinforcement Learning}

As discussed in section \ref{rw}, previously, adversary sample generation methods against Deep RL agents assume direct access to the inputs of the victim or its policy. In contrast, finding an adversarial policy with only interacting with a victim environment has been achieved for vision based agents in PvP environments such as simulated robotic games \cite{gleave2019adversarial}.

In this approach, instead of adding a perturbation to the victims input the adversary interacts with the same environment containing its victim trading agent. By embedding the victim in the environment from adversary's point of view, the attack is treated as a single agent RL problem.  Training goal for the adversary is to learn actions that changes the victim actions, minimizing its reward $R_{victim}(s_t,a_{victim},s_{t+1})$ accumulated throughout the trading episode. These actions may seem unintuitive from the human prospective.

Although finding an adversarial policy for simulation games following a deterministic model is quite different from a trading environment with uncertainty and volatility, we employ our version of this method to find an trading adversary agent that is able to alter victim agent decisions for the worse.

\section{ADVERSARIAL POLICY FOR ATTACKING TRADING AGENTS}\label{appr}

\subsection{Adversary Policy}
In this research, we aim to demonstrate a gray-box approach for attacking a deep reinforcement learning trading agent, since the main exchange systems used by traders are very safe and almost unreachable from outside, meaning there is no simple way of manipulating data received by trading algorithms. We also assume no access to the trading agents source code, input, policy network architecture and training algorithm. Our only viable data would be current state of the environment and the decision of the trading agent or it's chosen action in that given state. Adversary agent is provided with the combination of inputs to trading agent's DNN policy and output of the agent. Our proposed adversary agent uses a DNN as its policy. This DNN consists of convolutional and fully connected layers as used in most computer vision tasks. The convolutional part in the networks captures a more appropriate representation of temporal information from the data as well as existing relations between different features. It is also an effective way of canceling out noises appearing in the data similar to noisy pixels in images. Overall, this method helps to increase decision certainty of the fully connected layers of DNN compared to the raw data points. We use Categorical Cross Entropy loss to train the adversary using only 4 days of stock market data (8\% of test data). 
Given the intrinsic temporal dependencies present in stock market data, the application of RNNs may constitute a more apt approach. However, in this study, we chose a less complex architectural design in contrast to competing trading agents. This choice is made with the primary objective of elucidating the apparent influence of the adversary.

\subsection{Reward Function}\label{rewfunc}
The reward function must represent the trading task and maximize the returns, while reducing the certainty of the trading policy decisions. 
It also need to be easily optimized. 
We propose $R$ as reward function for our adversary agent:  
\begin{equation}
 R = (Balance + P-\hat{P}) \times \alpha + \vert\pi(a\vert S) - \pi(\hat{a}\vert\hat{S})\vert
\end{equation}

$Balance$ is the amount of currency available at the agents disposal at each step. $P$ the value of agents portfolio or the value of its owned shares and $\hat{P}$ is the changed portfolio value after performing action $a$ by the adversary. $\alpha$ is a scaling factor determined in the training process. $\pi$ is the victims policy making its trading decisions give states $S$ and $\hat{S}$.
Scaling assets in $R$ by $\alpha$ encourages the adversary to have more emphasis on changing the trading agents' decisions hence, not over fitting on other components of reward function such as its own cumulative returns. To maintain the purpose of feasibility in our proposed approach in a real-world setting, we assume a soft constraint on the money spent by adversary agent by giving it a fixed budget at the start of trading.
Another important constraint that should not be overlooked by the agent while making buy orders is market liquidity. Market liquidity is total amount of stocks available to buy in the market. The agent should be able to detect if there are no available shares to buy using a given state vector from trading environment. We will describe state vector in Section \ref{encode}.

Finally, since the adversary agent's decisions are trades in a market that can generate profit or cause losses, changing the decision of the trading agent is not a indicator of adversary's performance by itself.  To address this issue, Adversary's Reward function takes asset loss caused by the changes in traders decision into account, as well.

\subsection{Advantage Actor Critic (A2C)}
An actor-critic based algorithm in reinforcement learning is a policy gradient algorithm that tries to find an approximation of the value function as well as a policy at the same time. Value function is a prediction of future rewards given the current state of an agent. This tells the agent how good a state is for it to be in. 
Arbitrary fluctuations in price, volume of trades and other features of the trading market data means it is of stochastic nature with unknown transitions to the agent. To efficiently train the proposed adversary we use A2C, or Advantage Actor Critic algorithm. A2C is a deterministic and synchronous implementation of A3C\cite{mnih2016asynchronous}.

A2C benefits from an ensemble technique or the advantage function, reducing policy gradient variance for each update resulting in a more robust policy. This method gathers multiple gradient updates from different instances of the same policy but using different data points 
During each iteration A2C averages over all of the  calculated gradients by different instances and updates the actor and critic networks accordingly. More general gradient updates improves speed and rate model convergence, making this algorithm suitable for the problem of trading by reducing the effect of noisy or uncertain actions.


\subsection{Real-time Trading Environment}
\subsubsection{Trading Simulation}
\begin{figure}
\centerline{\includegraphics[width=0.5\textwidth]{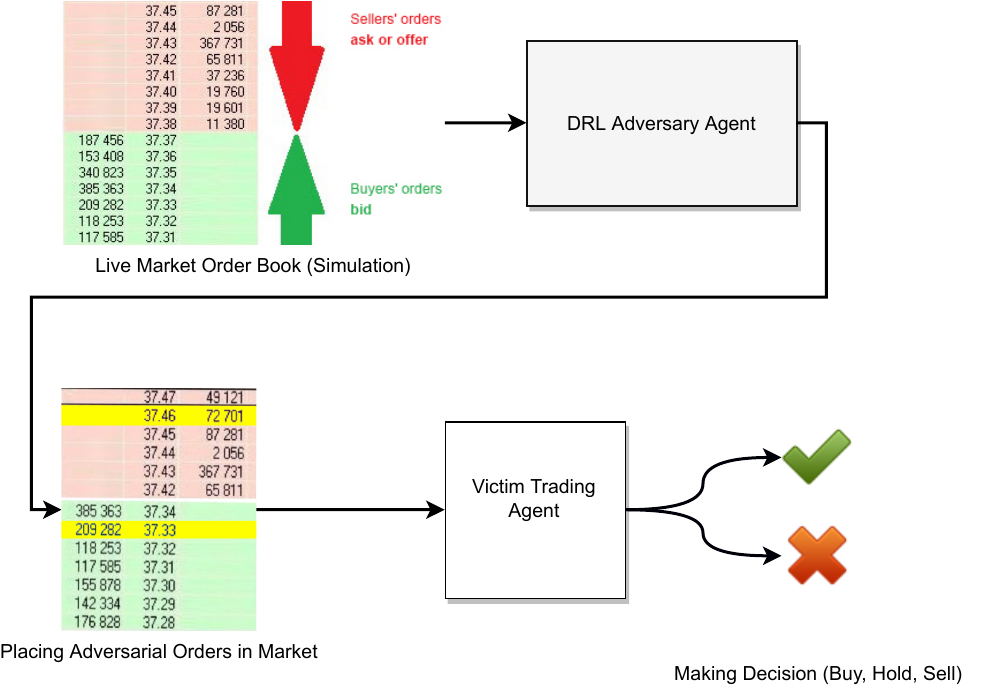}}
\caption{Overview of Limit Order Book or LOB impacted by adversarial attacks in the trading environment architecture}
\label{adrl}
\end{figure}

Having a dynamic trading market data that reacts according to agents' decisions is a crucial part of our research. We want the agents' orders to have real-time impact on environment to be practical as much as possible. Therefore we have chosen ABIDES which is an agent-based trading market simulator that provides a trading data with very similar latent space to the real market as shown in their experiments \cite{10.1145/3384441.3395986}. 
ABIDES provides an API for agents to place orders, cancel or modify them at their desired timing. Then it uses an exchange agent for market making. 
As we use OpenAI Gym's implementations for our experiments, we have integrated ABIDES into the Gym environment. At each time step, we gather the full Limit Order Book (LOB) from the simulation as it represents market state in the most accurate and detailed way. Trading agents are then provided with top-10 bids and asks appending useful indicators extracted from LOB to decide weather to place an order or not (Figure \ref{adrl}). 

\subsubsection{Policy Input Encoding}\label{encode}
The bids ($bid_i$) from buyers and asks ($ask_j$) from sellers that currently exist on the market are maintained in the simulation ordered from best to worst. Each $bid_i$ and $ask_j$ are prices corresponding to a buyer or seller agent. The simulator knows each agent from its id shown as $agent_{i}$ and $agent_{j}$. List of bids and asks can be presented as ordered lists of tuples:

\begin{equation}
\begin{split}
    bids = \langle (bid_i,agent_{i})&, (bid_{i+1}, agent_{i+1}), ...  \rangle  \\ & , bid_i > bid_{i+1} \\  \\
    asks = \langle (ask_j,agent_{j})&, (ask_{j+1}, agent_{j+1}), ...  \rangle \\  & , ask_j > ask_{j+1}
\end{split}
\end{equation}

We generate an input vector as State($S_t$) of the trading environment at time $t$. Elements of this vector are calculated using historical stock price, asks and bids vectors collected over time. We define state vector as:
\begin{equation*}
V_{t} = 
\begin{bmatrix} 
	B ,	V , asks , n_{asks} , bids , n_{bids} ,RSI , CCI , MACD 
\end{bmatrix} 
\end{equation*}
Where:
\begin{itemize}
    \item{ $B\in\mathbb{R}$: Is the remaining currency balance available to the agent at a given time step.}
    \item $V\in\mathbb{R}$: Is the number of shares in agents wallet bought in previous time steps.
    \item $asks \in \mathbb{R}^{10}$: 10 best asking prices in the market at time t.
    \item $n_{asks} \in \mathbb{N}^{10}$: Amount of available shares to buy at each asking price.
    \item $bids \in \mathbb{R}^{10}$: 10 best biding prices in the market at time t
    \item $n_{bids} \in \mathbb{N}^{10}$: Amount of available shares to buy at each biding price
    \item $RSI \in \mathbb{R}$: Relative Strength Index is calculated from collected stock prices. RSI is a technical indicator that helps traders to analyze recent momentum of a stock and measure whether a stock is oversold or undersold in a trading market. \cite{chong2014revisiting}.
    \item $CCI \in \mathbb{R}$: Commodity Channel Index is also calculated from collected stock prices. CCI is a technical indicator known for its proficiency in detecting cyclical trends in stock markets \cite{maitah2016commodity}.
    \item $MACD \in \mathbb{R}$: MACD is a momentum indicator that shows the relationship between two moving averages of price and is a well-known trend following technical indicator used to analyse stock markets \cite{chong2014revisiting}.
\end{itemize}

This vector will be recalculated and fed directly to adversary's neural network based policy by the training environment in each time step.

\subsubsection{Training Environment}
OpenAI Gym provides a perfect framework for training a vast range of agents on different datasets and simulations. However, it does not offer an environment for agents to bet against each other. We start by training the trading agent using our environment and save the best performing policy checkpoints. In the next step, we train the adversary in a simulated environment where at each time step both agents in play are provided with $S_{t_i}$ 
Adversary agent is also provided with output of the trading policy or $a$ and the profit made from this single action to make a decision to place an order. Based on the adversary decision, we update $S_{t_i}$ to measure the impact of market change caused by adversary. 


\section{Experimental Evaluation}\label{evl}
Our reserach objective can be addressed with these RQs:
\textbf{ RQ1: How effective is the proposed adversary in changing the trading algorithms' decisions?} 
To elaborate, in RQ1 we ignore the actual profit loss of the trading agent caused by the adversary and focus only on the trading agent's Softmax policy output. To answer this question we compare the policy output of the trading agent before and after updating the simulation with orders from adversary policy. 

We also measured victims' rewards for the natural and under attack actions taken by them. The trading agents in this study are able to use a reward function that represents the quality of their actions in terms of returns during their training. We took advantage of these preexisting functions to collect our data.

\textbf{RQ2: To what extent the adversary algorithm is able to change trader’s profits?}
As mentioned, in section \ref{rewfunc}, trader's policy outputs will be treated as trades in the market. Each of these trades can cause loss or generate profit, based on the stock price changes. However, only changing trader's decision in one step does not guarantee a trend in its profit/loss. The trader may be able to compensate for the losses of a single trade (or even become profitable after several consecutive steps) by changing its next decisions. 

\begin{figure}
\centerline{\includegraphics[width=0.5\textwidth]{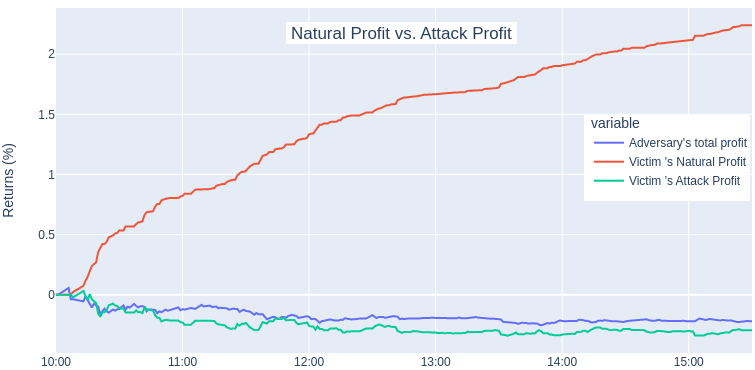}}
\caption{Returns of the industrial trader in a sample episode and returns of the same agent in the same episode while under attack by the proposed adversary.} 
\label{tva.png}
\end{figure}

In a close to real-world trading scenario, a successful attack performed by an adversary should compel the trader to lose profit in the market by changing its decision. To measure the effects of adversarial attacks on trader, we run the same market simulation twice in parallel. Once without the adversary in play, and a second time while adversary is attacking the trader by placing orders. (see Figure \ref{tva.png})

\textbf{RQ3: How well does the proposed algorithm able to maximize trading agent's portfolio gain or loss with maintaining reasonable constraints? And is the adversary exploiting any specific patterns of trading to attack victims?}
One of the key identifiers of the proposed adversary's efficacy is the amount of sacrificed resources. The adversary might be able to manipulate the trader, however, it should be able to do this by maintaining a feasible loss margin for itself, while trading. That is it should not consume an unforeseeable amount of its budget and impose only a little profit damages to itself. We answer this RQ by tracking adversary's assets consisting of its balance and bought shares.
Furthermore, we dive deeper into details of our proposed adversary to explore its trading methods in order to change its victim's decision. Our proposed adversary is trained against three types of trading victims and is tasked with learning strategies fit for attacking these specific type of traders. To study adversarial agents' trading behaviour against each victim, we also track adversary's episode rewards in parallel with it's direct trades with a victim to gain an insight on the strategies of placing adversarial market orders.

\subsection{Evaluation Metrics}
\subsubsection{RQ1 Performance Metrics}
We seek to measure the severity of changes in trader's behavior in each state. The first metric we report is average change in Softmax output of trader's policy network. The original state of simulation, where there is no attacker placing orders, is denoted as $S_t$ and the state when attacker is present is denoted as $\hat{S_t}$. The trader policy's average output change for an episode of $N$ steps is defined as follow:
\begin{equation}
\label{rq11}
\Delta_{episode} = \frac{1}{N} \sum_{t=1}^{N}\big\vert \frac{\pi(a_t\vert S_t) - \pi(\hat{a_t}\vert\hat{S_t})}{\pi(a_t\vert S_t)}\big\vert \times 100
\end{equation}

As the second efficiency metric for the adversary method, we define average rewards over $N$ steps of an episode as:
\begin{equation}
\label{rq12}
\bar R = \frac{1}{N} \sum_{t=1}^{N} R_{a_t}^{S_t}
\end{equation}

We report natural rewards (no attack) alongside reward under adversarial attacks for over 50 episodes for 3 trading agents.
To evaluate the difference in the reward distributions of natural and attack rewards, we use a goodness of fit test for each victim.


Since collected reward data from our experiments belong to continuous distributions ($R \sim D_{R}$) and includes 50 data points for each experiment, Kolmogorov-Smirnov (KS) statistical test has been chosen to measure the distance between distribution of natural rewards ($D_{R_{natural}}$) and distribution of attack rewards $D_{R_{attack}}$. KS test is non-parametric and distribution-free, meaning it makes no assumption over the distribution of data. The KS test can be used to compare a sample with a reference probability distribution, or to compare two samples. Null hypothesis for KS test is that these two distributions are identical and come from the same distribution($D$), for all data points; the alternative is that they are not identical, in case of rejection of the null hypothesis. 
\begin{equation}
\label{null}
    R_{natural},R_{attack} \stackrel{i.i.d}{\sim} D
\end{equation}
  
 \begin{equation*}
    R_{natural},R_{attack} \stackrel{i.i.d}{\nsim} D
\end{equation*} 

\subsubsection{RQ2 Performance Metrics}
In RQ2, we evaluate the impact of changes in trading environment states by adversary on trader's portfolio. Two metrics are used as follows:
\begin{itemize}
    \item Cumulative reduction in returns per episode: Given $P$ is trader's returns without the presence of the adversary and $\hat{P}$ is trader's returns under attack, it is calculated as:
    \begin{equation}
    \label{rq21}
    CR = \frac{1}{N_e}\sum_{t=1}^{N_e}\frac{P-\hat{P}}{P} \times 100
    \end{equation}
    We measure average of changes in cumulative returns over $N_e = 50$ episodes.
    
    \item Average reduction in returns per step: Given $p$ is trader's returns in an individual step without the presence of adversary and $\hat{p}$ is trader's returns while under attack, we define AOR as:
    \begin{equation}
    \label{rq22}
    AOR = \frac{1}{N}\sum_{t=1}^{N}\frac{{p-\hat{p}}}{p} \times 100
    \end{equation}
    We report best AOR for all of the $50$ episodes with $N$ number of steps to understand how the portfolio losses are distributed over single episodes.
\end{itemize}
\subsubsection{RQ3 Performance Metrics}
This RQ explains the behavior of the adversarial agent. For evaluating costs of the adversary policy trading in the market we report portfolio value returns of the attacker and compare it to the same metric $CR$ for the victim from RQ2. Here a successfull attacker should have a smaller loss compared to the victim, otherwise the attack would be too costly to worth it, in most scenarios.

For the second part of this RQ, which seeks answer to whether the attacker simply trades directly with the victim or has managed to disrupt it learning process more systematically, we report two performance metrics from adversary. The first metric is Mean Episode Rewards of the adversary gathered from OpenAI Gym. For the second metric, we introduce Loss Hit-Ratio. Essentially, Loss Hit-Ratio is intended to measure what ratio of victim's losses are caused by directly trading with the adversary agent. In order to define loss hit-ratio, we maintain arrays of the agent ids to keep track of each bid and ask present in the trading environment at any given time:

\begin{equation*}
\begin{split}
Bid_{ids} = & [(AGENT_1, shares_1,price_1),\\ & ... , (AGENT_{10}, shares_{10},price_{10})] 
\end{split}
\end{equation*}

\begin{equation*}
\begin{split}
Ask_{ids} = & [(AGENT_1, shares_1,price_1),\\ & ... , (AGENT_{10}, shares_{10},price_{10})]
\end{split}
\end{equation*}

These vectors tell us that which bid and ask belong to which agent. 
Therefore, we can exactly calculate the profit or loss gained from a specific share bought from an order placed by the adversary. Given a completed exchange, if the victim sells or buys shares directly from the adversary the exchange is counted as a hit. Thus the Loss Hit Ratio for the victim is:

\begin{equation*} 
     \text{Loss Hit Ratio} = \dfrac{\text{Returns from hits}}{\text{Total returns}}
\end{equation*}

A successful attack is expected to have a low Loss Hit Ratio to show that the victim is not simply trading only with the adversary, as explained above.

\subsection{Real-time Data Generation (Simulation)}
We ran the market simulation embedded in training environment over 50 times for each experiment using 3 distinct market configurations provided by its developers. These configurations use a specific number of different trading agents such as noise agents and momentum agents including a single market maker as well as an exchange agent for handling orders (the details of different configurations can be found in the provided replication package). Each of the episodes start at 9:30am when the market opens and ends at 4pm. The trading and adversary agents under study collect price data and existing orders in LOB every 20ms which is the exact moment when exchange agent wakes up to organize existing orders. They are also allowed to place orders at the same moment since our experiments should mimic a real-time trading market.


\subsection{Experimental Setup}
Training and evaluation of each Deep RL agents for trader and adversary was done on a single machine running Ubuntu 20.04.2 LTS (Linux 5.8.0) equipped
with Intel Core i7-9700 CPU, 32 gigabytes of main memory, and 8
gigabytes of GPU memory on a NVIDIA GeForce RTX 2080 graphics
card. Implementation is done with Pytorch and OpenAI Gym.


\subsection{Results}
\subsubsection{RQ1 results}
Table \ref{tab:rq11-results} shows $\bar R$ and $\Delta_{episode}$ for three trading agents in 2 scenarios: 1) with the adversary in the environment (attack $\bar R$) and 2) without the adversary (natural $\bar R$) from equation \ref{rq12}. The highest value of both metrics that we have observed is included as well.

\begin{table*}
\tiny
\caption{Mean and best rewards and $\Delta$, over 50 episodes, for agents under attack and the natural reward (no attack)}
\label{tab:rq11-results}
\resizebox{\textwidth}{!}{%

\begin{tabular}{c|c|c|c|c|c|c}

 \textbf{\textbf{Trader Model}} & 
\textbf{Market} &
\textbf{Natural $\bar R$}  & \textbf{Attack $\bar R$} & \textbf{Best Attack $\bar R$}  & \textbf{$\Delta_{episode}$} & \textbf{Best $\Delta_{episode}$}

   \\
    
\multirow{3}{*}{\textbf{Baseline}}  & config1 & 0.541   & -0.484          & -2.031 & 46.9\%          & 51.2\% \\
  & config2 & 0.318   & -1.093          & -1.594 & 47.3\%          & 52.0\%\\ 
  & config3 & 0.332   & -1.146          & -2.309 & 39.8\%          & 44.8\% \\ 
\multirow{3}{*}{\textbf{Ensemble}}  & config1  & 0.727   &   -0.051  & -0.994 & 30.2\%          & 34.4\%\\
  & config2  & 0.611 & -0.823     & -2.062 & 27.4\%   & 32.1\% \\  
  & config3 & 0.698   & -0.983          & -2.137 & 28.5\%          & 29.6\% \\ 
\multirow{3}{*}{\textbf{Industrial}} & config1  & 0.598   & -0.291          & -1.003 & 22.1\%  & 36.4\% \\
 & config2 & 0.919 & -1.094          & -1.875 & 16.2\%     & 17.9\% \\ 
 & 
 config3 & 0.859   & -0.432   & -0.976 & 25.3\%           & 27.7\% \\

\end{tabular}%
}
\end{table*}

The first observation from the results is that reward function of the traders are showing a considerable amount of negative impact caused by the adversary. All of the trading algorithms show a positive mean reward ($\bar{R}$) in the trading environment, meaning their decisions at first are generating orders with acceptable returns in the course of an episode or trading day. But the mean reward received by the trading agent after making decisions under attacks, shows that our proposed adversary was able to force the victim to make incorrect trades and has impaired the ability of the trading agent to make a reliable prediction of the future stock price. Although the trading agents is provided with the same technical indicators, they are still vulnerable to seeing adversary orders in the LOB.

Looking at the $\Delta$ measurements in Table \ref{tab:rq11-results} we see a wide range (from 16.2\% to 47.3\%). The overall pattern is as expected: the baseline is easier to fool, then the ensemble method, and the Industrial model is the hardest to manipulate. 
However, we see that even the smaller manipulations of the policy's Softmax output (e.g., 16.2\% in the Industrial-Config2 case) can result in large declines in the reward values (from 0.919 to -1.094 in this example). 

Reported distance between the Natural and the Attack reward distributions in table \ref{tab:rq11-ks} shows a considerable difference in victims performance, while under attack. Since all of the p-values are extremely smaller than 0.05 which means the distance between $D(R_{natural})$ and $D(R_{attack})$ are calculated with confidence, we can safely claim that the defined Null hypotheses from equation (\ref{null}) has been rejected. 

\begin{table}
\centering
\tiny
\caption{Distances (and p-values) of Natural $\bar{R}$ and Attack $\bar{R}$}
\label{tab:rq11-ks}
\resizebox{\columnwidth}{!}{%

\begin{tabular}{c|c|c}

 \textbf{Trader Model} &
\textbf{Kolmogorov-Smirnov Distance} &
\textbf{p-value}    \\
    
 \textbf{Baseline} &  0.72 & 8.7593e-13
 \\ 
\textbf{Ensemble} &  0.64 & 6.0786e-10
 \\ 
 \textbf{Industrial} & 0.54 & 4.9291e-07
 \\

\end{tabular}%
}
\end{table}

To sum up RQ1, the average Natural $\bar R$ over all 9 trader-config pairs is 0.623 and the average Attack $\bar R$ is -0.711, which shows a (0.623-(-0.711))/0.623 = 214.17\% reduction in the reward value. 
This shows the effectiveness of our proposed adversary in forcing the agent to make non-optimal trades in the market, which are reflected in its reward function. 

\subsubsection{RQ2 results}
In RQ2, we report $CR$ and and $AOR$ from equations \ref{rq21} and \ref{rq22}. Both metrics are measured for various settings of environment, against different trading algorithms similar to RQ1. The results are presented in Table \ref{tab:rq2-results}.

\begin{table}
\tiny
\caption{$CR$ and $AOR$ for the baseline, the ensemble, and the industrial trading algorithms, averaged over 50 episodes.}
\label{tab:rq2-results}
\resizebox{\columnwidth}{!}{%

\begin{tabular}{c|c|c|c|c}

 \textbf{\textbf{Trader Model}} &
\textbf{Market} &
\textbf{CR}  & \textbf{AOR} & \textbf{Best CR}  \\
    
\multirow{3}{*}{\textbf{Baseline}}  & config1 & 82.88\%   & 118.9\%          & 88.30\% \\
  & config2 & 95.36\%   & 164.1\%          & 101.43\% \\ 
  & config3 & 85.59\%   & 135.2\%          & 90.09\%     \\ 
\multirow{3}{*}{\textbf{Ensemble}}  & config1  & 74.13\%   & 90.5\%          & 75.81\%    \\
  & config2  & 75.33\% & 97.8\%          & 77.64\%  \\ 
  & config3 & 73.92\%   & 92.7\%          & 77.72\% \\ 
\multirow{3}{*}{\textbf{Industrial}} & config1  & 71.05\%   & 97.9\%          & 81.14\%   \\
 & config2 & 63.68\% & 73.5\%          & 74.84\%  \\ 
 & config3 & 65.74\%   & 85.1\%          & 69.93\% \\

\end{tabular}%
}
\end{table}

Considering the $CR$s reported in the experiment, we can see that the proposed adversary is able to target victims' returns, by manipulating their trade decision effectively. Results show that the adversary not only is able to predict its victim's decision boundary (RQ1), but also learns to predict the trend of market price (represented as the returns and their reductions), by integrating a good representation of the market and victims trading strategy (RQ2). This makes our method efficient in generating targeted attacks (on profits) against trading agents as well as un-targeted attacks (only altering victims output). 

Looking at the reported $AOR$s, it is clear that our adversary is able to force the victim into making trading decisions that tend to work against the market trend. It reduces even the best trading agent's returns not only in the course of trading, but also in individual steps. $AOR$ gives us a better understanding of intensity of attacks, where they were able to reduce immediate profits (on average over the three market configs per trader) by 139.4\% for baseline trader, 93.7\% for ensemble, and 85.5\% for industry trader in its weakest attack.

\subsubsection{RQ3 results} 

To answer this RQ, we first look at the losses of the victims ($CR$) vs. the adversary's (Adversary Portfolio Loss), in table \ref{tab:rq3-results}. We can see that the adversary is able to reach its goal by spending a small percentage of its starting budget (100\% loss would mean using all the assigned budget to be able to fool the trader -- The initial budget of the adversary is set equal to the victim to have a fair comparison). 

\begin{table*}

\caption{Portfolio loss per episode and normalized mean episode rewards for the adversary compared to Victims' portfolio loss $CR$.}

\label{tab:rq3-results}
\resizebox{\textwidth}{!}{%

\tiny
\begin{tabular}{c|c|c|c|c|c}

\textbf{Trader Model} &
 \textbf{Market} &
\textbf{Victim $CR$}  & \textbf{Adversary Portfolio Loss} &  \textbf{Mean Episode Rewards} & \textbf{Loss Hit Ratio}\\
    
\multirow{3}{*}{\textbf{Baseline}}  & config1
& 82.88\%   &  16.04\%  & 0.8931 &  17.932\% \\
  & config2 
  & 95.36\%   & 13.82\%    & 0.9434 & 12.146\%   \\ 
 & config3 
  & 85.59\%   & 20.13\%     & 0.9789 & 16.753\% \\ 
\multirow{3}{*}{\textbf{Ensemble}}  & config1 
& 74.13\%   & 28.54\%   & 0.9103 &   14.301\% \\
  & config2  
  & 75.33\% & 27.76\%      & 0.9520 &    14.166\%   \\ 
  & config3 
  & 73.92\%   & 21.49\%      & 0.8447 &  15.353\%   \\ 
\multirow{3}{*}{\textbf{Industrial}} & config1  
& 71.05\%   & 37.66\%   & 0.8939 &  10.099\% \\
 & config2 
 & 63.68\% & 39.23\%  & 0.8942 &  13.993\% \\ 
 & config3 
 & 65.74\%   & 43.17\%  & 0.9007 & 13.067\% \\

\end{tabular}%
}
\end{table*}

Looking at example results against the baseline victim, our adversary (on average over the three market configs) had to consume $(87.94/16.66)-1 = 427.77\%$ less budget compared to its victim and $(74.46/25.93)-1 = 187.16\%$ less budget against the ensemble trading victim and $(66.82/40.02)-1 = 66.97\%$ less against the best trading victim. The table also shows that although the adversary has to place larger and possibly more trades subject to negative returns in the market in order to manipulate better trading victims, but even with the better victims, it was still able to reach its preferable outcome with less budget compared to the victim.

To have a better insight on how the adversary operates and analyze its learned strategy, we have represented mean episode rewards of the adversary alongside loss hit ratio in table \ref{tab:rq3-results}. Note that the rewards are relatively high in all of the experiments against victims even where the adversary has performed worse, which means the value function of the agent perceives the adversary trades to be efficient enough. 

Another interesting finding that is verified by loss hit ratio is that a small percentage of victims' loss is caused by directly trading with the adversary, which is an indicator of the adversary's strategy to disrupt the natural trading course of the victim. By combining two observations of high rewards and low loss hit ratio, we conclude that the adversary has learned a wining strategy. That is, rather than interacting with the victim directly, in most scenarios, it changes the limit order book to a more out-of-distribution observation compared to the training observation that victim is more familiar with.

\section{Related Work}\label{rw}

Previous studies on adversary sample generation for DNNs mostly focuses on directly modifying the inputs. Some studies found that deep neural networks are prone to mis-classification by adding perturbation undetectable by human vision to the input \cite{szegedy2013intriguing}. Furthermore, they show that these examples can be generalized over a variety of DNN architectures and training sets \cite{papernot2017practical}. Later studies introduced Fast Gradient Sign Method or FGSM \cite{goodfellow2014explaining}. FGSM exploits gradients of the DNN, approximating the model to generate adversarial examples. 

An early study on applications of adversary example generation using FGSM was done on several deep reinforcement learning algorithms (DQN\cite{van2016deep}, A3C\cite{mnih2016asynchronous}, TRPO\cite{schulman2015trust}). They found that FGSM is able to decrease the agents policy regardless of the environment, architecture and training algorithm. This method was applied in a white-box manner to generate the FGSM perturbation. Then they proceeded to use transferability of adversarial examples to attack RL agents in a black-box manner with only access to the DNN structure and training environment \cite{huang2017adversarial}.

Gradient based adversarial example generation methods have been studied for RL application in the trading domain as well \cite{chen2021adversarial, faghan2020adversarial}. Both of these methods attack the input channel of the victim directly and use historical stock exchange datasets. However, these assumptions renders both approaches non-feasible for the real-world trading scenario.  

A universal adversarial perturbations threat model was introduced by studying vulnerability of RL by generating fake orders in the stock market dataset \cite{goldblum2021adversarial}. They apply the perturbations to the test dataset by reiterating over all orders. This approach is still assuming a low-level access to the inputs by making custom changes to entries of the trading dataset.

In an interesting study, authors benchmarks collision avoidance ability of autonomous driving agents \cite{behzadan2019adversarial}. Their approach tests the robustness of RL agent behaviours in environments where they interact with other agents. Trading in an stock market is very similar to such environments specially zero-sum games where money lost by an agent is another agent's profit. In addition, some studies showed RL agents trained in collaboration or against other agents might get closely dependent on them and fail against different agents \cite{lanctot2017unified}. We deal with this issue by using numerous noise agents in the stock exchange simulation that is used to train the victims.

\section{Conclusion and Future Work}\label{conc}
This paper introduces a Deep RL adversary trading agent that can be used to test the lower-bound of trading agents in a very close to real-world stock market scenario. The proposed approach also shows that despite using complex deep neural networks in a trading agent's policy, they are still prone to natural, but out of distribution attacks by an adversary. We tested our approach on three different settings for market simulation against three different trading agents. 


Some potential extensions to this work include: (a) using the adversary to generate a defence method against such threats and (b) to train anomaly detection methods to alert automated trading agent or even the exchanges of such possible risks.

\bibliographystyle{IEEEtran}
\bibliography{citations}

\begin{thebibliography}{10}
\providecommand{\url}[1]{#1}
\csname url@samestyle\endcsname
\providecommand{\newblock}{\relax}
\providecommand{\bibinfo}[2]{#2}
\providecommand{\BIBentrySTDinterwordspacing}{\spaceskip=0pt\relax}
\providecommand{\BIBentryALTinterwordstretchfactor}{4}
\providecommand{\BIBentryALTinterwordspacing}{\spaceskip=\fontdimen2\font plus
\BIBentryALTinterwordstretchfactor\fontdimen3\font minus \fontdimen4\font\relax}
\providecommand{\BIBforeignlanguage}[2]{{%
\expandafter\ifx\csname l@#1\endcsname\relax
\typeout{** WARNING: IEEEtran.bst: No hyphenation pattern has been}%
\typeout{** loaded for the language `#1'. Using the pattern for}%
\typeout{** the default language instead.}%
\else
\language=\csname l@#1\endcsname
\fi
#2}}
\providecommand{\BIBdecl}{\relax}
\BIBdecl

\bibitem{mnih2013playing}
V.~Mnih, K.~Kavukcuoglu, D.~Silver, A.~Graves, I.~Antonoglou, D.~Wierstra, and M.~Riedmiller, ``Playing atari with deep reinforcement learning,'' \emph{arXiv preprint arXiv:1312.5602}, 2013.

\bibitem{silver2016mastering}
D.~Silver, A.~Huang, C.~J. Maddison, A.~Guez, L.~Sifre, G.~Van Den~Driessche, J.~Schrittwieser, I.~Antonoglou, V.~Panneershelvam, M.~Lanctot \emph{et~al.}, ``Mastering the game of go with deep neural networks and tree search,'' \emph{nature}, vol. 529, no. 7587, pp. 484--489, 2016.

\bibitem{dosovitskiy2017carla}
A.~Dosovitskiy, G.~Ros, F.~Codevilla, A.~Lopez, and V.~Koltun, ``Carla: An open urban driving simulator,'' in \emph{Conference on robot learning}.\hskip 1em plus 0.5em minus 0.4em\relax PMLR, 2017, pp. 1--16.

\bibitem{noonan2017jpmorgan}
L.~Noonan, ``Jpmorgan develops robot to execute trades,'' \emph{Financial Times}, pp. 1928--1937, 2017.

\bibitem{yang2020deep}
H.~Yang, X.-Y. Liu, S.~Zhong, and A.~Walid, ``Deep reinforcement learning for automated stock trading: An ensemble strategy,'' in \emph{Proceedings of the First ACM International Conference on AI in Finance}, 2020, pp. 1--8.

\bibitem{roa2019adversarial}
J.~Roa-Vicens, Y.~Wang, V.~Mison, Y.~Gal, and R.~Silva, ``Adversarial recovery of agent rewards from latent spaces of the limit order book,'' \emph{arXiv preprint arXiv:1912.04242}, 2019.

\bibitem{szegedy2013intriguing}
C.~Szegedy, W.~Zaremba, I.~Sutskever, J.~Bruna, D.~Erhan, I.~Goodfellow, and R.~Fergus, ``Intriguing properties of neural networks,'' \emph{arXiv preprint arXiv:1312.6199}, 2013.

\bibitem{huang2017adversarial}
S.~Huang, N.~Papernot, I.~Goodfellow, Y.~Duan, and P.~Abbeel, ``Adversarial attacks on neural network policies,'' \emph{arXiv preprint arXiv:1702.02284}, 2017.

\bibitem{gleave2019adversarial}
A.~Gleave, M.~Dennis, C.~Wild, N.~Kant, S.~Levine, and S.~Russell, ``Adversarial policies: Attacking deep reinforcement learning,'' \emph{arXiv preprint arXiv:1905.10615}, 2019.

\bibitem{10.1145/3384441.3395986}
\BIBentryALTinterwordspacing
D.~Byrd, M.~Hybinette, and T.~H. Balch, ``Abides: Towards high-fidelity multi-agent market simulation,'' in \emph{Association for Computing Machinery}, ser. SIGSIM-PADS '20, New York, NY, USA, 2020, p. 11–22. [Online]. Available: \url{https://doi.org/10.1145/3384441.3395986}
\BIBentrySTDinterwordspacing

\bibitem{repo}
\BIBentryALTinterwordspacing
 [Online]. Available: \url{https://anonymous.4open.science/r/ADRL-B72D/README.md}
\BIBentrySTDinterwordspacing

\bibitem{fischer2018reinforcement}
T.~G. Fischer, ``Reinforcement learning in financial markets-a survey,'' FAU Discussion Papers in Economics, Tech. Rep., 2018.

\bibitem{lillicrap2015continuous}
T.~P. Lillicrap, J.~J. Hunt, A.~Pritzel, N.~Heess, T.~Erez, Y.~Tassa, D.~Silver, and D.~Wierstra, ``Continuous control with deep reinforcement learning,'' \emph{arXiv preprint arXiv:1509.02971}, 2015.

\bibitem{mnih2016asynchronous}
V.~Mnih, A.~P. Badia, M.~Mirza, A.~Graves, T.~Lillicrap, T.~Harley, D.~Silver, and K.~Kavukcuoglu, ``Asynchronous methods for deep reinforcement learning,'' in \emph{International conference on machine learning}.\hskip 1em plus 0.5em minus 0.4em\relax PMLR, 2016, pp. 1928--1937.

\bibitem{chong2014revisiting}
T.~T.-L. Chong, W.-K. Ng, and V.~K.-S. Liew, ``Revisiting the performance of macd and rsi oscillators,'' \emph{Journal of risk and financial management}, vol.~7, no.~1, pp. 1--12, 2014.

\bibitem{maitah2016commodity}
M.~Maitah, P.~Prochazka, M.~Cermak, and K.~{\v{S}}r{\'e}dl, ``Commodity channel index: Evaluation of trading rule of agricultural commodities,'' \emph{International Journal of Economics and Financial Issues}, vol.~6, no.~1, pp. 176--178, 2016.

\bibitem{papernot2017practical}
N.~Papernot, P.~McDaniel, I.~Goodfellow, S.~Jha, Z.~B. Celik, and A.~Swami, ``Practical black-box attacks against machine learning,'' in \emph{Proceedings of the 2017 ACM on Asia conference on computer and communications security}, 2017, pp. 506--519.

\bibitem{goodfellow2014explaining}
I.~J. Goodfellow, J.~Shlens, and C.~Szegedy, ``Explaining and harnessing adversarial examples,'' \emph{arXiv preprint arXiv:1412.6572}, 2014.

\bibitem{van2016deep}
H.~Van~Hasselt, A.~Guez, and D.~Silver, ``Deep reinforcement learning with double q-learning,'' in \emph{Proceedings of the AAAI conference on artificial intelligence}, vol.~30, no.~1, 2016.

\bibitem{schulman2015trust}
J.~Schulman, S.~Levine, P.~Abbeel, M.~Jordan, and P.~Moritz, ``Trust region policy optimization,'' in \emph{International conference on machine learning}.\hskip 1em plus 0.5em minus 0.4em\relax PMLR, 2015, pp. 1889--1897.

\bibitem{chen2021adversarial}
Y.-Y. Chen, C.-T. Chen, C.-Y. Sang, Y.-C. Yang, and S.-H. Huang, ``Adversarial attacks against reinforcement learning-based portfolio management strategy,'' \emph{IEEE Access}, vol.~9, pp. 50\,667--50\,685, 2021.

\bibitem{faghan2020adversarial}
Y.~Faghan, N.~Piazza, V.~Behzadan, and A.~Fathi, ``Adversarial attacks on deep algorithmic trading policies,'' \emph{arXiv preprint arXiv:2010.11388}, 2020.

\bibitem{goldblum2021adversarial}
M.~Goldblum, A.~Schwarzschild, A.~Patel, and T.~Goldstein, ``Adversarial attacks on machine learning systems for high-frequency trading,'' in \emph{Proceedings of the Second ACM International Conference on AI in Finance}, 2021, pp. 1--9.

\bibitem{behzadan2019adversarial}
V.~Behzadan and A.~Munir, ``Adversarial reinforcement learning framework for benchmarking collision avoidance mechanisms in autonomous vehicles,'' \emph{IEEE Intelligent Transportation Systems Magazine}, vol.~13, no.~2, pp. 236--241, 2019.

\bibitem{lanctot2017unified}
M.~Lanctot, V.~Zambaldi, A.~Gruslys, A.~Lazaridou, K.~Tuyls, J.~P{\'e}rolat, D.~Silver, and T.~Graepel, ``A unified game-theoretic approach to multiagent reinforcement learning,'' \emph{Advances in neural information processing systems}, vol.~30, 2017.

\end{thebibliography}
\end{document}